\documentclass[letterpaper, 10 pt, journal, twoside]{IEEEtran}

\IEEEoverridecommandlockouts               
\usepackage{cite}
\usepackage{color,xcolor}
\usepackage{graphicx}
\usepackage{pdfpages}
\usepackage{subfigure}
\usepackage{tabulary}
\usepackage{balance}

\usepackage{amsmath}
\usepackage{amssymb}
\usepackage{float}
\usepackage{soul}
\usepackage{booktabs}
\usepackage{multirow}  
\usepackage{hyperref} 
\usepackage{array}  
\usepackage{diagbox}

\usepackage{mathtools}
\graphicspath{{media/}}
\usepackage{bm}
\usepackage{verbatim}

\usepackage[T1]{fontenc}

\begin{document}

\pagestyle{empty} 

\title{\LARGE Neural Autoencoder-Based Structure-Preserving Model Order Reduction and Control Design for High-Dimensional Physical Systems}
    
\author{
    Marco Lepri$^{1,*}$, Davide Bacciu$^2$, Cosimo Della Santina$^{3,4}$
    \thanks{This work is supported by the EU EIC project EMERGE, grant number 101070918. 
    $^1$NEC Laboratories Europe, Heidelberg, Germany. Email: marco.lepri@neclab.eu
    $^2$Università di Pisa, Italy. Email: davide.bacciu@unipi.it
    $^3$Department of Cognitive Robotics, Delft University of Technology, Building 34, Mekelweg 2, 2628 CD Delft, Netherlands. Email: c.dellasantina@tudelft.nl.
    $^4$Institute of Robotics and Mechatronics, German Aerospace Center (DLR), 82234 Wessling, Germany.
    $^*$Work partially done while the author was at University of Pisa}
}

\maketitle
\thispagestyle{empty} 

\begin{abstract} This work concerns control-oriented and structure-preserving learning of low-dimensional approximations of high-dimensional physical systems, with a focus on mechanical systems. 
We investigate the integration of neural autoencoders in model order reduction, while at the same time preserving Hamiltonian or Lagrangian structures. We focus on extensively evaluating the considered methodology by performing simulation and control experiments on large mass-spring-damper networks, with hundreds of states. The
empirical findings reveal that compressed latent dynamics with less than 5 degrees of freedom can accurately reconstruct the original systems' transient and steady-state behavior with a relative total error of around 4\%, while simultaneously accurately reconstructing the total energy. Leveraging this system compression technique, we introduce a model-based controller that exploits the mathematical structure of the compressed model to regulate the configuration of heavily underactuated mechanical systems.

 \end{abstract}

\IEEEpeerreviewmaketitle

\section{Introduction}

\IEEEPARstart{S}{everal} 
application domains exhibit high-dimensional dynamics, e.g., continuum mechanics, fluid dynamics, quantum systems, financial markets. In such contexts, a useful approach for effective control, which often relies on system-specific expertise, is to find low-dimensional approximations of these systems that preserve their key structural properties \cite{Agrachev2005,della2021model,Kaptanoglu2020}. This work concerns itself with automatic discovery of these approximations using machine learning.

In machine learning, a wealth of research focuses on approximating complex nonlinear dynamical systems while ensuring the learned dynamics fulfill specific structural properties \cite{Greydanus2019,Cranmer2020,beckers2022gaussian,evangelisti2022physically}, which enabled application to model-based control \cite{liu2023physics,lutter2023combining}.
The case of direct learning of a compressed dynamics of an high-dimensional system has also been thoroughly investigated in the literature and applied to model-based control
\cite{gillespie2018learning,Brunton2022,Perera2022,mahlknecht2022using}.

A relevant alternative to directly learn the dynamics combines analytic models with machine learning \cite{swischuk2019projection}. An established strategy is to project the dynamics into a latent space using principal component analysis (PCA) \cite{willcox2002balanced}. Nonlinear counterparts of PCA, such as neural autoencoders (AE), have been considered only in recent years in \cite{Fulton2018} and \cite{Shen2021}. These pioneering works target the rendering of deformable objects in computer graphics, only providing qualitative analysis of simulation behavior.

\begin{figure}[t]
    \centering
    \includegraphics[width=\linewidth]{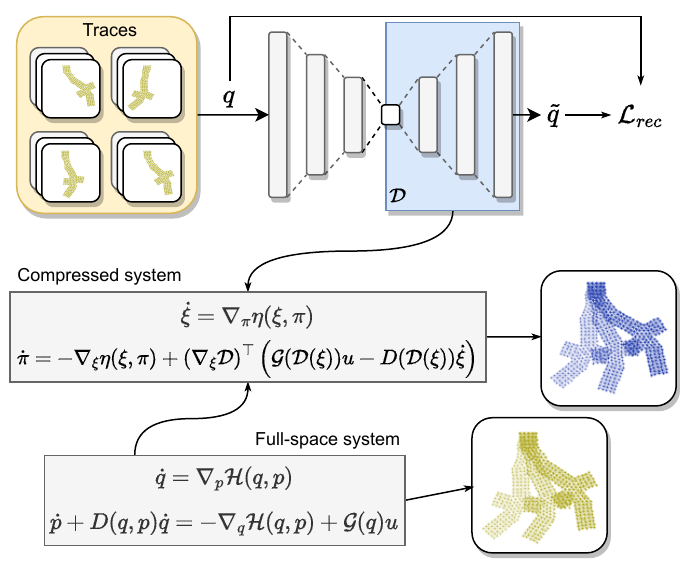}
    \label{fig:approach}
    \caption{\small The proposed strategy is a two-step process. First, we compress the configuration space $q$ of the physical system into latent representations via deep neural autoencoders. We then generate a compressed dynamical system that uses the learned latent representation while maintaining the Hamiltonian structure of the complete system.}
    \vspace{0.1cm}
\end{figure}
\begin{figure*}
    \centering
    \includegraphics[width=0.9\linewidth]{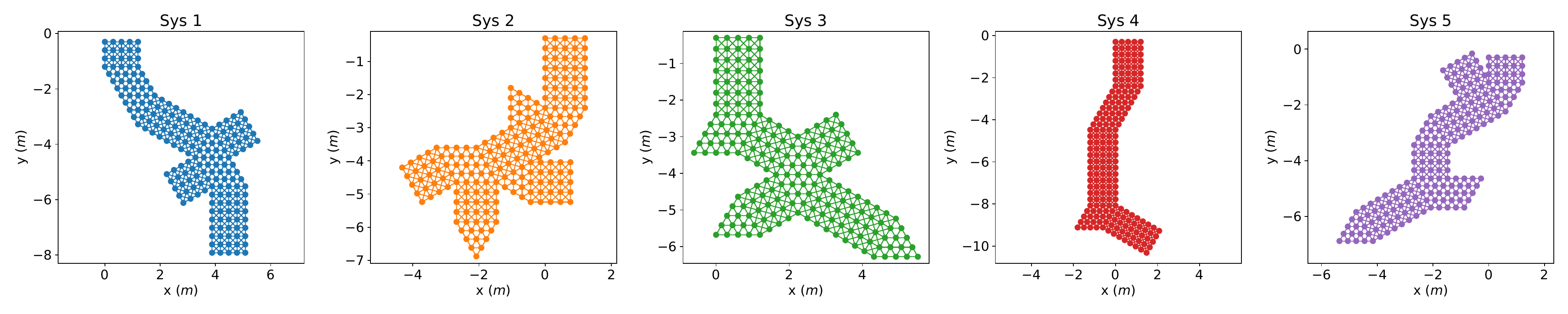}
    \caption{\small The five planar models of deformable objects considered in this works, in their rest position when no gravity is present. Each model is a mass-spring-damper network composed of 205 masses and 636 connections.}
    \label{fig:systems}
\end{figure*}

In this work, we make a further step in that direction by combining deep learning with structure-preserving model order reduction \cite{karasozen2021structure,hesthaven2021structure}. Our approach is schematized in Figure \ref{fig:approach}. We exploit AEs, investigating both flat \cite{Goodfellow2016} and graph-based AEs \cite{gentleIntro}, to extract compressed representations of the system's configuration directly from evolution traces. 
Then, by combining the decoder of the autoencoder model with the original system specification, we derive a new set of dynamic equations describing the system's dynamics, while at the same time maintaining their original Hamiltonian or Lagrangian form.
Relying on such a structure, we also propose a closed-loop controller that can regulate the configuration of high-dimensional systems relying on a small amount of inputs.
We thoroughly test these methodologies on networks of masses interconnected by springs and dampers that can be seen as a finite element approximation a continuous mechanical system.
We conduct numerical simulations in the latent space, focusing on the adherence of the reduced dynamics to the original system and its physical principles. In addition, we explore the capability of the reduced system to approximate the real system under highly constrained latent space dimensions. Finally, experiments on planar posture regulation are performed, exploiting the learned representations and developed controller.

This approach holds potential for diverse applications involving soft robots \cite{armanini2021soft} or deformable objects \cite{zhu2022challenges}. For instance, it could address control tasks on soft robots, leveraging approximations of their state to deal with their high dimensionality and limited actuation. Similarly, it may find application in the manipulation of deformable objects, where analogous limitations and challenges exist.

\section{Deep Physical Compression}
A generic Port-Hamiltonian\footnote{Analogous derivations would be possible in the Lagrangian one.} system is defined as 
\begin{equation}\small 
    \dot{x} = \left[\mathcal{J}(x) - \mathcal{R}(x)\right]\nabla_x\mathcal{H}(x) + \mathcal{G}(x)u,
\end{equation}
where $x$ is the system's state, and $u$ the input. $\mathcal{J}$ is a skew-symmetric matrix that specifies the interconnection structure, $\mathcal{R}$ is a semi-positive definite dissipation matrix, $\mathcal{G}$ is the input field, and $\mathcal{H}$ is the Hamiltonian of the system - i.e., its total energy.
We consider here systems whose state can be represented as $x = (q,p) \in \mathbb{R}^{2n}$, with $q$ being the configuration and $p$ the generalized momenta, and having the following structure
\begin{equation}\small \label{eq:system}
    \dot{q} = \nabla_{p}\mathcal{H}(q,p), \; \dot{p} = - D(q)\nabla_{p}\mathcal{H}(q,p) - \nabla_{q}\mathcal{H}(q,p) + \mathcal{G}(q)u,
\end{equation} 
where $D(q) \in \mathbb{R}^{n \times n}$ is a dissipation matrix, assumed positive definite. The term $D(q)\nabla_{p}\mathcal{H}(q,p)$ is a common way to describe effects that make the energy strictly decrease in time, as mechanical friction. Note indeed that $\dot{\mathcal{H}} = - \dot{q}^{\top}D(q)\dot{q} \leq 0$. The control action $u$ is assumed to be of size $a$ and thus $\mathcal{G} \in \mathbb{R}^{n \times a}$.
For the sake of clarity of derivations, we assume the Hamiltonian $\mathcal{H}:\mathbb{R}^{2n} \rightarrow \mathbb{R}$ to be quadratic in the generalized momenta
\begin{equation}\small \label{eq:energy}
    \mathcal{H}(q,p)=\dfrac{1}{2}p^{\top}M^{-1}(q)p+V(q),
\end{equation}
where $M:\mathbb{R}^{n} \rightarrow \mathbb{R}^{n\times n}$ is a positive definite matrix and $V:\mathbb{R}^{n} \rightarrow \mathbb{R}$ the potential energy. Note that $\dot{\mathcal{H}} = -\left(M p\right)^{\top} D \left(M p\right) \leq 0$ as $D \succeq 0$.
For example, mechanical systems have such a structure. In this case, $M$ is called the inertia matrix.\\
We assume that a description of the system in the form \eqref{eq:system} is available. Our goal is to obtain a new system with the same Hamiltonian structure but with a substantially smaller state space, leveraging the concepts described in the following.

\paragraph{Autoencoders}
We propose to use a neural autoencoder \cite{Goodfellow2016} to compress the configuration space representation from dimension $n$ to dimension $m << n$. An autoencoder is composed of two parts. 
\begin{itemize}
    \item An encoder network $\mathcal{E}:\mathbb{R}^{n} \rightarrow \mathbb{R}^{m}$ that compresses $q$ into its latent representation $\xi \in \mathbb{R}^{m << n}$,
    \item a decoder network $\mathcal{D}:\mathbb{R}^{m} \rightarrow \mathbb{R}^{n}$ which maps $\xi$ in an approximation of $q$.
\end{itemize}
An ideal autoencoder is one such that $\mathcal{E}\left(\mathcal{D}\right)$ is close to the identity function, despite $m << n$. Since we want to solely assess the robustness of the deep compressor, we use a simple MSE loss without task-specific regularizations
\begin{equation}\small \label{eq:loss}
    \mathcal{L}_{\mathrm{rec}}(q) = ||q - \mathcal{D}(\mathcal{E}(q))||_2^2.
\end{equation}

\paragraph{Compressed System}
We perform derivations by assuming an ideal autoencoder, i.e., one for which the loss in \eqref{eq:loss} is close to zero. We will discuss this hypothesis later.
We want to give to the latent dynamics the same Hamiltonian structure of the complete system \eqref{eq:system}-\eqref{eq:energy}. We thus impose the following latent dynamics
\begin{equation}\small \label{eq:system_red}
    \dot{\xi} = \nabla_{\pi}\eta(\xi,\pi), \quad \dot{\pi} + \Delta(\xi)\nabla_{\pi}\eta(\xi,\pi) = - \nabla_{\xi}\eta(\xi,\pi) + \Gamma(\xi)u,
\end{equation} 
with $\pi \in \mathbb{R}^{m << n}$ being the generalized momenta associated to the latent space configuration $\xi \in \mathbb{R}^{m}$ introduced in the previous subsection. The terms $\Delta, \Gamma$ describe the latent-space dissipation and input field respectively. The latent Hamiltonian/energy is
\begin{equation}\small \label{eq:energy_red}
    \eta(\xi,\pi)=\dfrac{1}{2}\pi^{\top}M_{\eta}^{-1}(\xi)\pi+V_{\eta}(\xi),
\end{equation}
with $M_{\eta}$ and $V_{\eta}$ being latent space counterparts of $M$ and $V$.
We now need to derive all the unknowns from the knowledge of the original dynamics and of the autoencoder. We start by relating the time derivative of the latent configuration with the one of the full configuration by the chain rule $\dot{q} = \nabla_{\xi} \mathcal{D} (\xi) \, \dot{\xi}$, where $\nabla_{\xi} \mathcal{D}$ is the Jacobian of the decoder. Combining the first equation in \eqref{eq:system_red} and \eqref{eq:energy_red}, we also get $\pi = M_{\eta}(\xi) \dot{\xi}$. We then impose that the latent energy is the same as the total energy of the system
\begin{equation}\small \label{eq:latent_energy}
    {\eta}(\xi,\pi) = \mathcal{H}(\mathcal{D} (\xi), p(\xi,\pi))
\end{equation}
where $p(\xi,\pi)$ is a mapping from the latent state to $p$, and $\mathcal{H}$ is defined as in \eqref{eq:energy}. The following choices of compressed potential $V_{\eta}$ and inertia matrix $M_{\eta}$ fulfill the constraints imposed by \eqref{eq:energy}, \eqref{eq:energy_red}, and \eqref{eq:latent_energy}
\begin{equation}\small \label{eq:potential_red}
    V_{\eta}(\xi) = V(\mathcal{D}(\xi)) 
\end{equation}
and
\begin{equation}\small \label{eq:kinetic_red}
    \begin{split}
        \pi^{\top} M_{\eta}^{-1}(\xi) \pi &= p^{\top} M^{-1}(\mathcal{D}(\xi)) p, \\
         \Rightarrow \; \dot{\xi}^{\top} M_{\eta} M_{\eta}^{-1} M_{\eta} \dot{\xi} &= \dot{\xi}^{\top}\left(\nabla_{\xi} \mathcal{D}\right)^{\top} M M^{-1} M \nabla_{\xi} \mathcal{D} \dot{\xi},\\
        \Rightarrow M_{\eta}(\xi) &= \left(\nabla_{\xi} \mathcal{D}(\xi)\right)^{\top} M(\mathcal{D}(\xi)) \; \nabla_{\xi} \mathcal{D}(\xi).
    \end{split}
\end{equation}
Comparing \eqref{eq:system} and \eqref{eq:system_red} and following similar steps as for the energy yields the following expressions for the input field and the dissipation
\begin{equation}\small \label{eq:input_diss_red}
    \Gamma(\xi) = \left(\nabla_{\xi} \mathcal{D}(\xi)\right)^{\top} \mathcal{G}(\mathcal{D}(\xi)), \, \Delta(\xi) = \left(\nabla_{\xi} \mathcal{D}(\xi)\right)^{\top} D(\mathcal{D}(\xi)).
\end{equation}
To conclude, combining together \eqref{eq:system_red}, \eqref{eq:energy_red}, \eqref{eq:potential_red}, \eqref{eq:kinetic_red}, and \eqref{eq:input_diss_red} yields the compressed Hamiltonian system in \eqref{eq:system_red_real}.

\begin{figure}
    \centering
    \includegraphics[width=\linewidth]{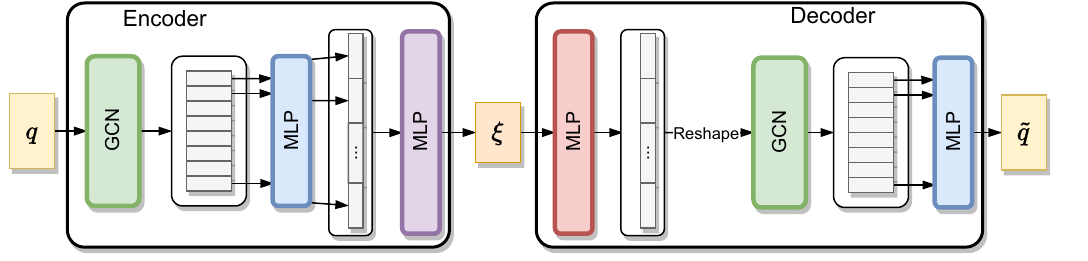}
    \caption{\small Graph Autoencoder architecture. Encoder and decoder are implemented as a combination of graph convolutional networks (GCN), to naturally process the graph, and multi-layer perceptrons (MLP), to process the obtained node embeddings.}
    \label{fig:gae}
\end{figure}
\begin{figure*}
\begin{equation}\small \label{eq:system_red_real}
    \begin{split}
        \dot{\xi} &= \nabla_{\pi}\eta(\xi,\pi), \qquad \dot{\pi} = - \nabla_{\xi}\eta(\xi,\pi) + \left(\nabla_{\xi} \mathcal{D}(\xi)\right)^{\top}\left(\mathcal{G}(\mathcal{D}(\xi))u - D(\mathcal{D}(\xi))\nabla_{\pi}\eta(\xi,\pi)\right), \\
    \text{with } \eta(\xi,\pi) &= \underbrace{\dfrac{1}{2}\pi^{\top}\left(\left(\nabla_{\xi} \mathcal{D}(\xi)\right)^{\top} M(\mathcal{D}(\xi)) \; \nabla_{\xi} \mathcal{D}(\xi)\right)^{-1} \pi}_{\text{Latent Space Kinetic Energy}}+\underbrace{V(\mathcal{D}(\xi))}_{\text{L.S. Potential E.}}.
    \end{split}
\end{equation}
\vspace{-0.5cm}
\begin{center}
    \rule{.6\textwidth}{0.25pt} 
\end{center}
\vspace{-0.4cm}
\end{figure*}

This is a low-dimensional dynamical system with the same mathematical structure of the original high-dimensional one \eqref{eq:system}. The two models will represent similar behaviors\footnote{Here, similar means in terms or real and reconstructed transients and steady-states, i.e., the error between the real and reconstructed transients and steady-states is reasonably low.} if $\mathcal{L}_{\mathrm{rec}} \simeq 0$.

\section{Learning Compressed Representations}
Compressed representations are obtained as the latent representations of a neural autoencoder whose architecture depends on the nature of the input information used to encode the uncompressed system. To show the flexible formulation of our approach, in the following we consider two alternative autoencoder configurations. The former is a flat autoencoder, comprising dense feed-forward layers in the encoder and decoder, where the uncompressed system in input is represented by the configuration vector $q$. The second is a graph autoencoder which leverages a structured representation of the physical systems meant to highlight their composing parts (e.g. masses) and the relationships existing between them (the adjacency constraints).

Deep learning for graphs (DLG) \cite{gentleIntro} deals with the adaptive processing of information represented in a structured form. These models typically work by learning to represent the structural elements (nodes, edges) or the full graphs in embedding vectors $\mathbf{h}$ which can then be used for predictive, descriptive, or generative purposes. The most popular DLG paradigm leverages a message passing scheme \cite{Gilmer2017} exploiting local information exchanges between neighboring nodes and exploits a layered neural architecture (where layering can be defined also by unfolding in time) to promote effective information diffusion across the graph. More formally, the encoding of the $v$-th node at layer $l+1$ is obtained as
\begin{equation}\small \label{eq:simple-aggregation}
    \mathbf{h}_v^{l+1} = \phi^{l+1} \Big(\mathbf{h}_v^l,\ \Psi(\{\psi^{l+1}(\mathbf{h}_u^{l}) \mid u \in \mathcal{N}_v\} ) \Big)
\end{equation}
where $\phi^{l+1}$, $\psi^{l+1}$ are parameterized neural layers (linear/nonlinear), and $\Psi$ is a permutation invariant function defined over the embeddings $\mathbf{h}_u^{l}$ of the nodes $u$ in the neighborhood $\mathcal{N}_v$ of $v$, computed at previous step $l$. 
The general formulation in equation \eqref{eq:simple-aggregation} can be specialized to cover a wide variety of DLG models, as shown in \cite{gentleIntro}. Within the scope of this work, we use a graph autoencoder with the architecture in Figure \ref{fig:gae} where both encoder and decoder are implemented with a specialization of \eqref{eq:simple-aggregation} using SAGE \cite{Hamilton2017} for neighborhood aggregation followed by an ELU non-linearity in $\phi$ (GCN block in the figure). The encoder obtains the latent embedding $\xi$ of the full graph through MLPs, which are also used in the decoder to reconstruct in output the node features $\tilde{q}$.

We comment here on the assumption, made in Section II-b, that the autoencoder achieves close-to-zero loss. In general, this is not trivial to achieve, nor to validate, for any given configuration, apart for those in the training set. However, the dissipative nature of the considered systems, guarantees that the set of \textit{reasonable} configurations is just a portion of the full configuration space. Therefore, it is much more reasonable to assume close-to-zero loss only on that subset, which can be more easily validated using dense enough simulated data as external validation/test set.

\section{Latent space control}
We consider under-actuated posture regulation - i.e., we want to generate a control action $u\in\mathbb{R}^a$ such that the high-dimensional configuration $q\in\mathbb{R}^n$ of system \eqref{eq:system} reaches $\bar{q}\in\mathbb{R}^n$, with $a<n$. 
Call $\bar{\xi} = \mathcal{E}(\bar{q}) \in \mathbb{R}^m$ the compressed encoding of $\bar{q}$ and assume that the system state $(q,\dot{q})$ is compressed online into $(\xi,\dot{\xi})$ through $\mathcal{E}$ and its Jacobian. The controller we propose has the form:
\begin{equation}\small 
    u(\bar{\xi}, \xi, \dot{\xi}) = A^{\mathrm{L}}(\bar{\xi}) \left(\underbrace{\frac{\partial V(\mathcal{D}(\xi))}{\partial \xi}(\bar{\xi})}_{\text{FeedForward}} + \underbrace{\alpha(\bar{\xi} - \xi) - \beta\pi}_{\text{Feedback}}\right),
    \label{eq:controller}
\end{equation}
where $A = \left(\nabla_{\xi} \mathcal{D}(\xi)\right)^{\top}\mathcal{G}(\mathcal{D}(\xi))$, with $A^{\mathrm{L}}$ its left inverse, and $\alpha, \beta \in \mathbb{R}^{+}$ are positive control gains.
This controller is essentially operating an output regulation when taking $\mathcal{E}(x)$ as output. The task-space closed loop generated by \eqref{eq:system_red_real} and \eqref{eq:controller} is 
$\dot{\pi} = \left[(\nabla_{\xi}\eta(\xi,\pi) - \nabla_{\xi}\eta(\bar{\xi},0)) + \alpha(\bar{\xi} - \xi)\right] - \left[\left(\nabla_{\xi} \mathcal{D}(\xi)\right)^{\top}\left(D(\mathcal{D}(\xi))\nabla_{\pi}\eta(\xi,\pi)\right) + \beta\pi\right]$, where we used that $A A^{\mathrm{L}} = I$ and that $\nabla_{\xi}\eta(\bar{\xi},0) = \nabla_{\xi}\left(V\circ\mathcal{D}\right)(\bar{\xi})$.
The convergence follows with standard arguments that we do not report here for the sake of space, under the assumption that $\alpha I + (\nabla_{\xi}\eta(\xi,\pi) - \nabla_{\xi}\eta(\bar{\xi},0))$ is positive definite in $\bar{\xi},0$. The closed loop energy $\eta(\xi,\pi) + \alpha(\bar{\xi} - \xi)^{\top}(\bar{\xi} - \xi)$ can be used as Lyapunov candidate, which has time derivative $\dot{V} = \pi^{\top}[\nabla_{\xi} \mathcal{D}(\xi)^{\top}D(\mathcal{D}(\xi))\nabla_{\xi} \mathcal{D}(\xi) + \beta]\pi \leq 0$, and invoking La Salle principle. In turn, convergence to the task space equilibrium implies $||\mathcal{E}(q_t) - \mathcal{E}(\bar{q})|| \rightarrow 0$.

Note that we could leverage this common arguments \cite{ortega2021pid} to design the controller and sketch a convergence proof because our learning technique is such that the task-space dynamics of $\mathcal{E}(x)$ is an Hamiltonian system.

\section{Simulations}

\subsection{Setup}
We evaluate the performance of the approach in compressing high-dimensional mechanical systems. We focus on continuously deformable planar bodies subject to a gravitational field and generic external perturbing forces. This class of systems is quite relevant from an application perspective as it is central in robotic manipulation of deformable objects \cite{zhu2022challenges} and control of soft robots \cite{della2021model}. 
We consider mass-spring-damper models as high-dimensional models of these systems \cite{armanini2021soft}. This is a widely used technique to approximate soft-body dynamics by discretizing their volume as a set of masses (nodes) interconnected by ideal springs and dampers (edges). We use simulation data of the systems to train an autoencoder to reconstruct their configuration $q$. Then, exploiting the learnt latent representation $\xi$, we simulate the corresponding full-space system by solving the compressed dynamics equation \eqref{eq:system_red_real}.

\begin{figure}
    \centering
    \includegraphics[width=0.95\linewidth]{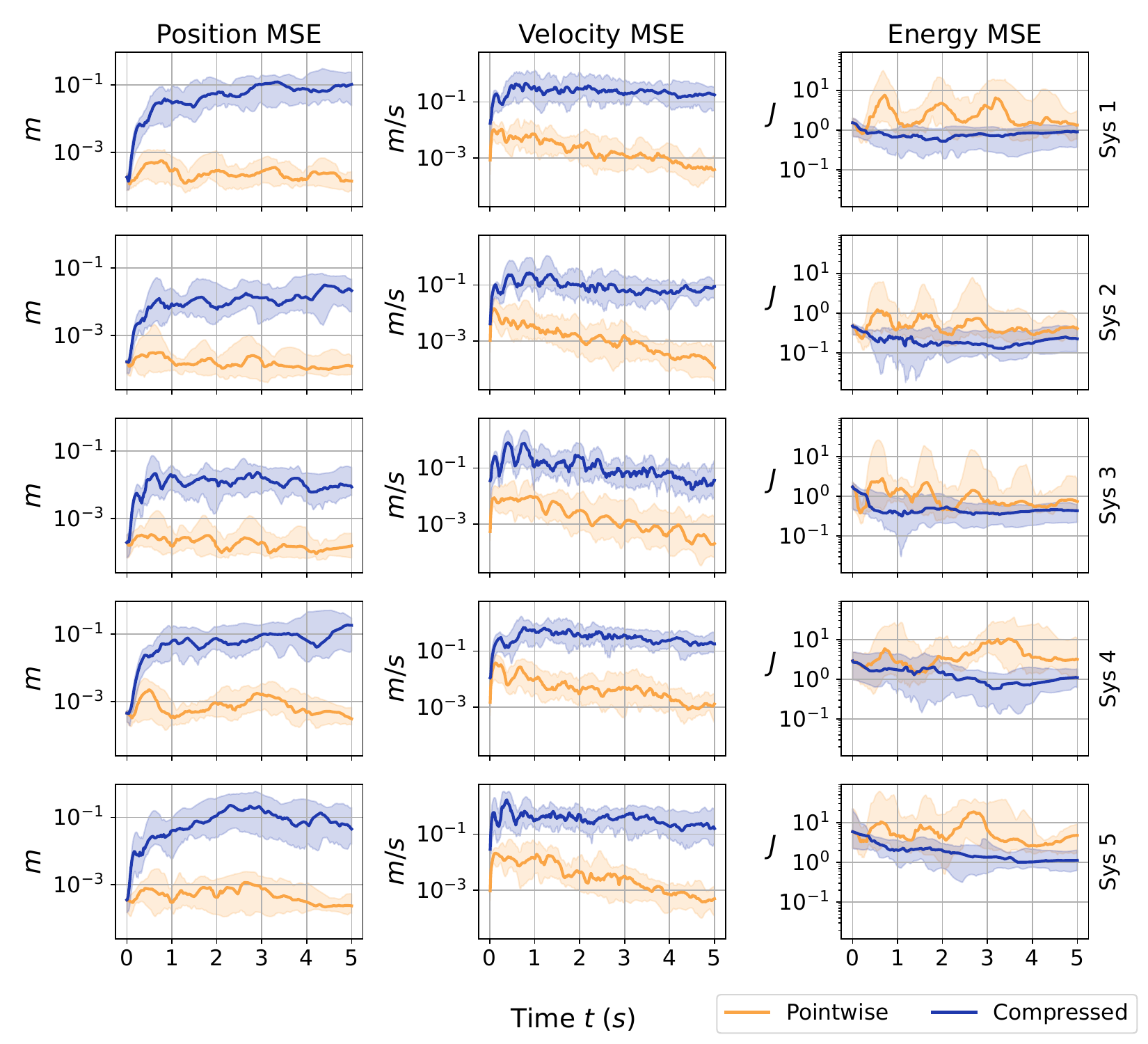}
    \includegraphics[width=0.95\linewidth]{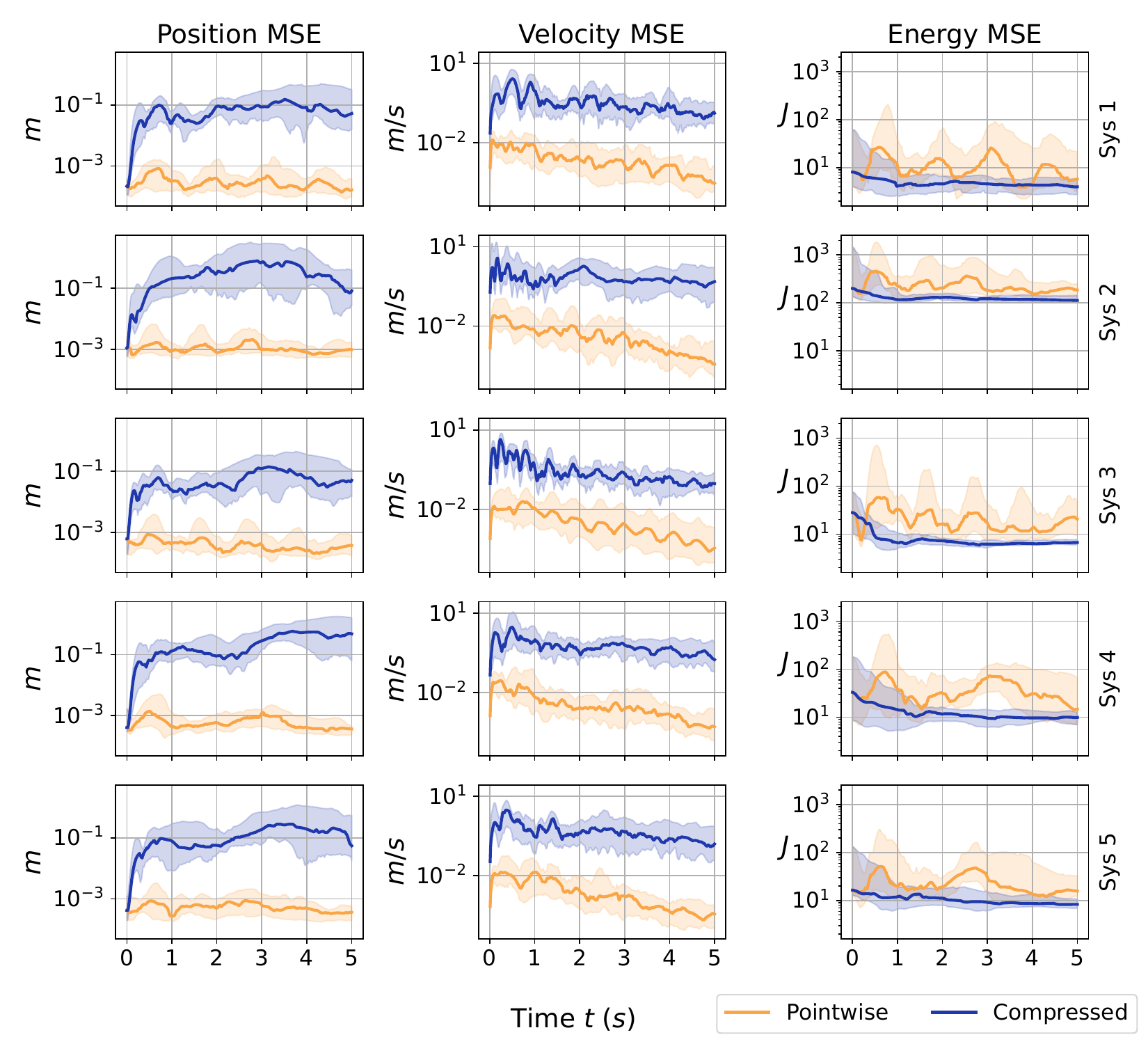}
    \caption{\small MSE on the reconstructed position $\mathcal{D}(\xi)$, velocity $(\nabla_\xi\mathcal{D})\dot{\xi}$ and total energy $\eta(\xi,\pi)$ of the systems in the test trajectories using: a flat autoencoder (left) and a graph autoencoder (right). For each plot, the bold line is the median error over the trajectories, while the shaded area represents the 20-80 percentiles.}
    \label{fig:mse}
\end{figure}

\paragraph{Data}
We use five high-dimensional, randomly generated systems in the form of \eqref{eq:system}. Figure \ref{fig:systems} shows the considered systems in their rest position. Each system is made of 200 masses and $e=636$ connections resulting in $n=400$ degrees of freedom captured in the configuration vector $q$. We consider systems immersed in a constant gravitational field. For each system, we perform 7 simulations to generate training data and 28 simulations for test purposes. Gravity conditions change for each simulation. The initial configuration $q(0)$ is randomly generated, while the systems always start at rest, i.e., $p(0) = 0$. More details on the data generation can be found in Section A.1 of the supplementary material.
\begin{table}[b]
\centering
\caption{Training, validation and test MSE score for the best flat and graph autoencoders, for each system.}
\begin{tabular}{c|c|c|c|c}
    \toprule
    System & Model & Train. mse & Valid. mse & Test mse \\
    \midrule
    \multirow{2}{*}{1} & Flat AE  & $2.17\mathrm{e}{-4}$  & $4.73\mathrm{e}{-4}$  & $\mathbf{4.24\boldsymbol{\mathrm{e}}{-4}}$ \\
                       & Graph AE & $3.40\mathrm{e}{-4}$  & $5.24\mathrm{e}{-4}$  & $7.25\mathrm{e}{-4}$ \\
    \midrule
    \multirow{2}{*}{2} & Flat AE  & $1.13\mathrm{e}{-4}$  & $2.48\mathrm{e}{-4}$  & $\mathbf{3.08\boldsymbol{\mathrm{e}}{-4}}$ \\
                       & Graph AE & $12.24\mathrm{e}{-4}$ & $15.37\mathrm{e}{-4}$ & $17.94\mathrm{e}{-4}$ \\
    \midrule
    \multirow{2}{*}{3} & Flat AE  & $1.69\mathrm{e}{-4}$  & $4.41\mathrm{e}{-4}$  & $\mathbf{3.75\boldsymbol{\mathrm{e}}{-4}}$ \\
                       & Graph AE & $3.73\mathrm{e}{-4}$  & $8.13\mathrm{e}{-4}$  & $7.37\mathrm{e}{-4}$ \\
    \midrule
    \multirow{2}{*}{4} & Flat AE  & $4.75\mathrm{e}{-4}$  & $18.08\mathrm{e}{-4}$ & $12.51\mathrm{e}{-4}$ \\
                       & Graph AE & $5.56\mathrm{e}{-4}$  & $13.72\mathrm{e}{-4}$ & $\mathbf{9.71\boldsymbol{\mathrm{e}}{-4}}$ \\
    \midrule
    \multirow{2}{*}{5} & Flat AE  & $8.53\mathrm{e}{-4}$  & $11.7\mathrm{e}{-4}$  & $12.70\mathrm{e}{-4}$ \\
                       & Graph AE & $7.41\mathrm{e}{-4}$  & $9.48\mathrm{e}{-4}$  &
                       $\mathbf{9.59\boldsymbol{\mathrm{e}}{-4}}$ \\
    \bottomrule
\end{tabular}
\label{tab:scores}
\end{table}

\paragraph{Training and Model Selection.}
A flat autoencoder and a graph autoencoder are trained for each system, with a latent size of 5 units. The same neural architecture is used across systems, while hyperparameters are selected via grid search for each system. Training, validation and test reconstruction scores (MSE) of the best model, for each system, are reported in Table \ref{tab:scores}.
More details on the training and model selection can be found in Section A.2 of the supplementary material.

\subsection{Simulation Results}

\begin{figure}
    \centering
    \includegraphics[width=0.95\linewidth]{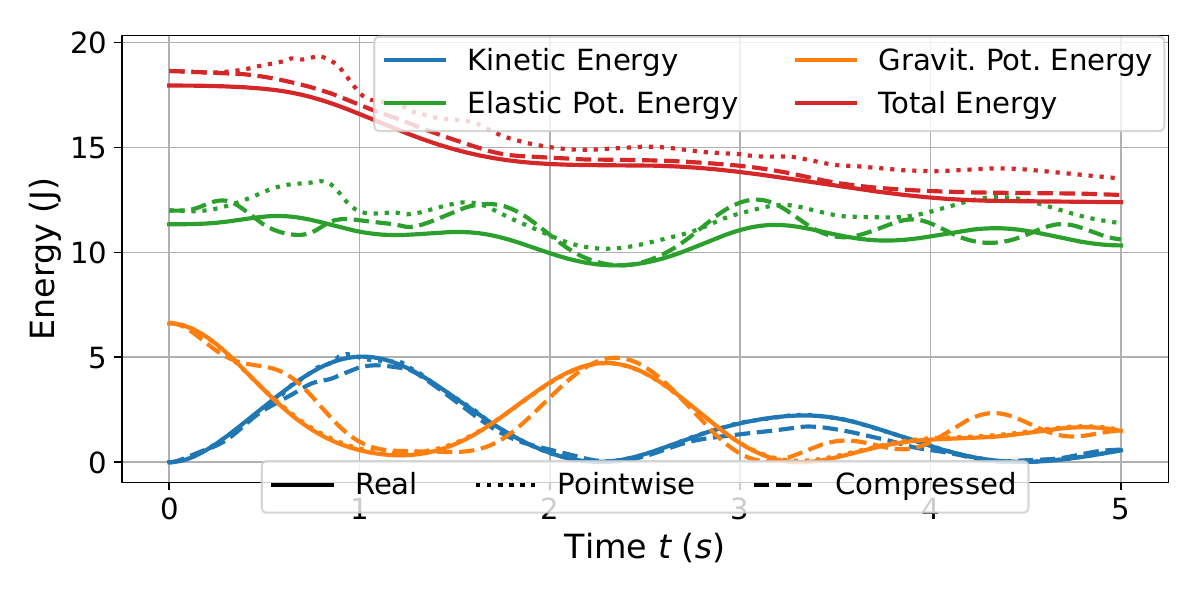}
    \includegraphics[width=0.95\linewidth]{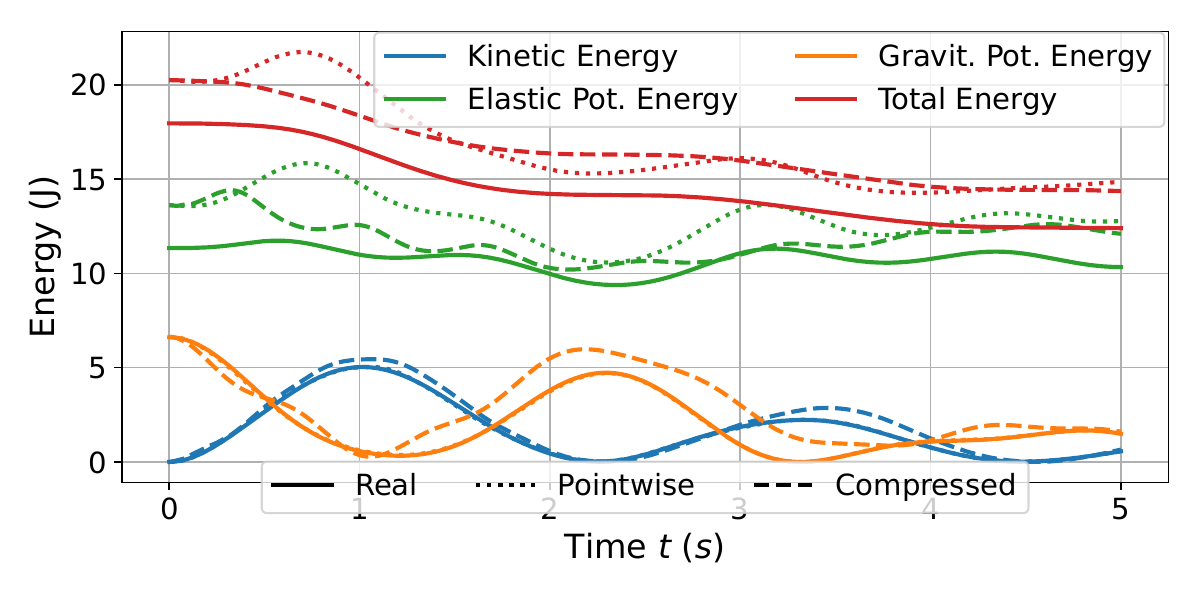}
    \vspace{-0.2cm}
    \caption{\small Example of evolutions of the kinetic, potential and total energies in a simulation for the full-space and reduced systems using: a flat autoencoder (top) and a graph autoencoder (bottom). The total energies are the Hamiltonians $\mathcal{H}$ and $\eta$  introduced in \eqref{eq:energy} and \eqref{eq:latent_energy} respectively. The solid line is the energy of system \eqref{eq:system}, while the dotted and dashed lines are the pointwise reconstructed and compressed energy.}
    \label{fig:energy}
\end{figure}

\begin{figure}
    \includegraphics[width=\linewidth]{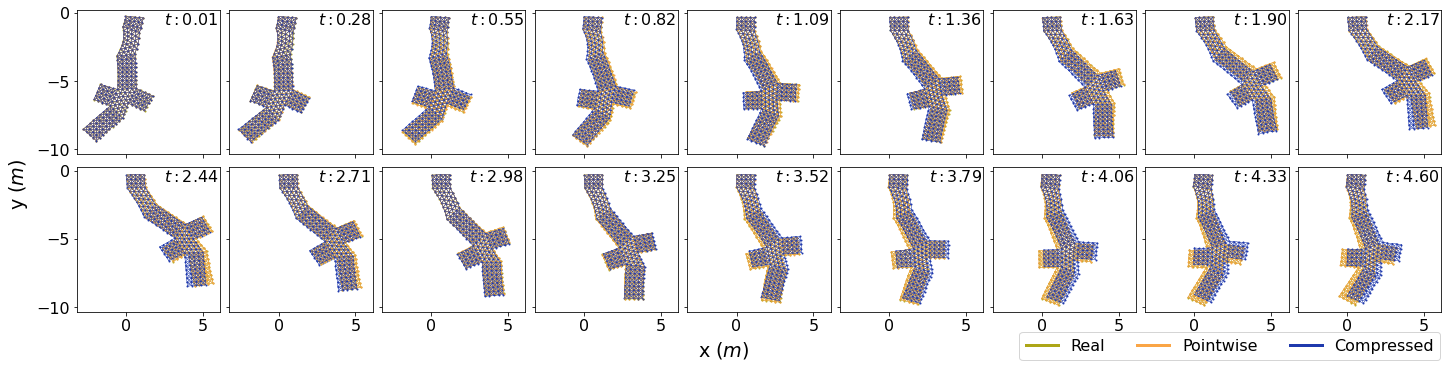}
    \includegraphics[width=\linewidth]{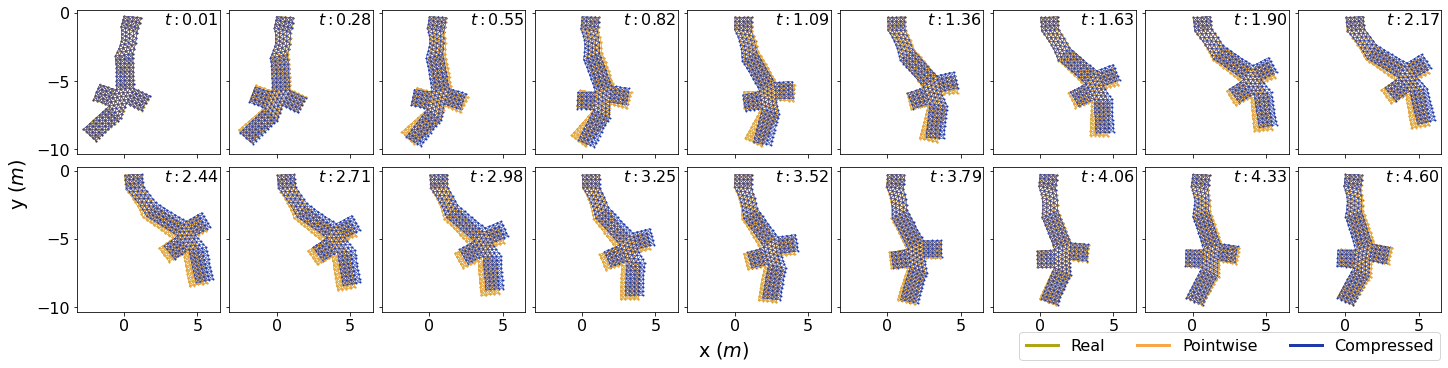}
    \caption{\small Comparison between frames from a real simulation (yellow), a pointwise reconstructed simulation (orange), and a reduced simulation (blue) using a flat autoencoder (above) and a graph autoencoder (below).}
    \label{fig:frames}
\end{figure}

In order to validate and evaluate the approach, we use the trained models to reconstruct the test full-space simulations, and we compare results in terms of reconstructed state $(q, \dot{q})$ and energy $\mathcal{H}$. In particular, we evaluate the approach in two different ways. First, for {\it pointwise evaluation}, we reconstruct the system state and energy at each step of the real simulation by applying the autoencoder end-to-end to the real system configuration $\mathcal{D}(\mathcal{E}(q))$ and velocity $\nabla_{\mathcal{E}(q)}\mathcal{D} \,\cdot \nabla_q\mathcal{E} \, \dot{q}$. Second, we use equation \eqref{eq:system_red_real} to simulate the evolution in the compressed space and reconstruct the configuration $\mathcal{D}(\xi)$, velocity $(\nabla_\xi\mathcal{D})\dot{\xi}$ and energy $\eta(\xi,\pi)$ from the latent variables at each simulation step ({\it compressed evaluation}).

Figure \ref{fig:mse} shows the average pointwise and compressed MSE w.r.t. time for each system. The two models have similar results in the compressed simulations, with the graph autoencoder having a slightly higher error, although it largely depends on the considered system. In all the cases, the compressed error on $q$ is a few orders of magnitude higher than the pointwise error. 
Interestingly, the error on $\dot{q}$ follows a similar behaviour although the models are never explicitly trained to reconstruct this information.

The reconstructed energy $\eta$ has the opposite trend: it is better reconstructed in the compressed case than in the pointwise case. This is due to the fact that pointwise reconstructing the state does not allow energy variations to govern the dynamics of the system, while this is possible in the compressed case. 
As a further evidence, Figure \ref{fig:energy} shows the evolution of the kinetic, potential, and total energy in a simulation. While the single kinds of energy might not be reconstructed as precisely in the compressed case, the total energy maintains the same non-increasing behaviour typical of dissipative systems.
Qualitatively, compressed simulations are stable and loyal to the real ones, with a good match of the full transient. Figure \ref{fig:frames} shows some frames from an example simulation. The portions where the real and reconstructed systems are not perfectly aligned are also those that exhibit major and more varied oscillations, such as the bottom portion of the central chain or some lateral structures.
Videos can be found \href{https://github.com/marcomerton/Deep-Physical-Compressor}{here}.

\subsection{Compression Analysis}
\begin{figure}
    \centering
    \includegraphics[width=\linewidth]{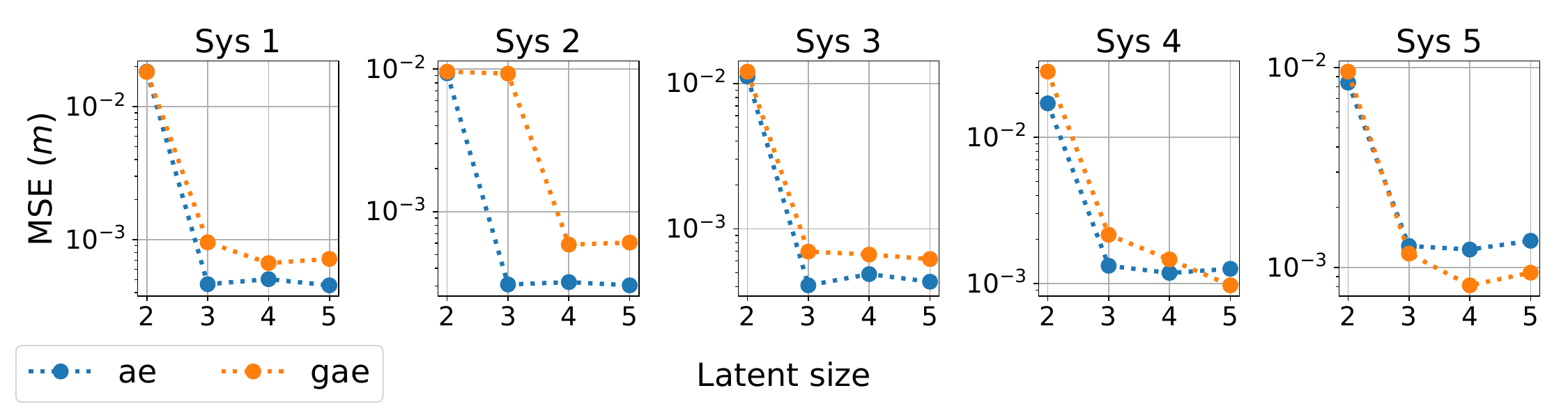}
    \caption{\small Test MSE on the five considered systems varying the latent state size of a flat autoencoder and a graph autoencoder.}
    \label{fig:compression}
\end{figure}

\begin{figure}
    \centering
    \includegraphics[width=0.9\linewidth]{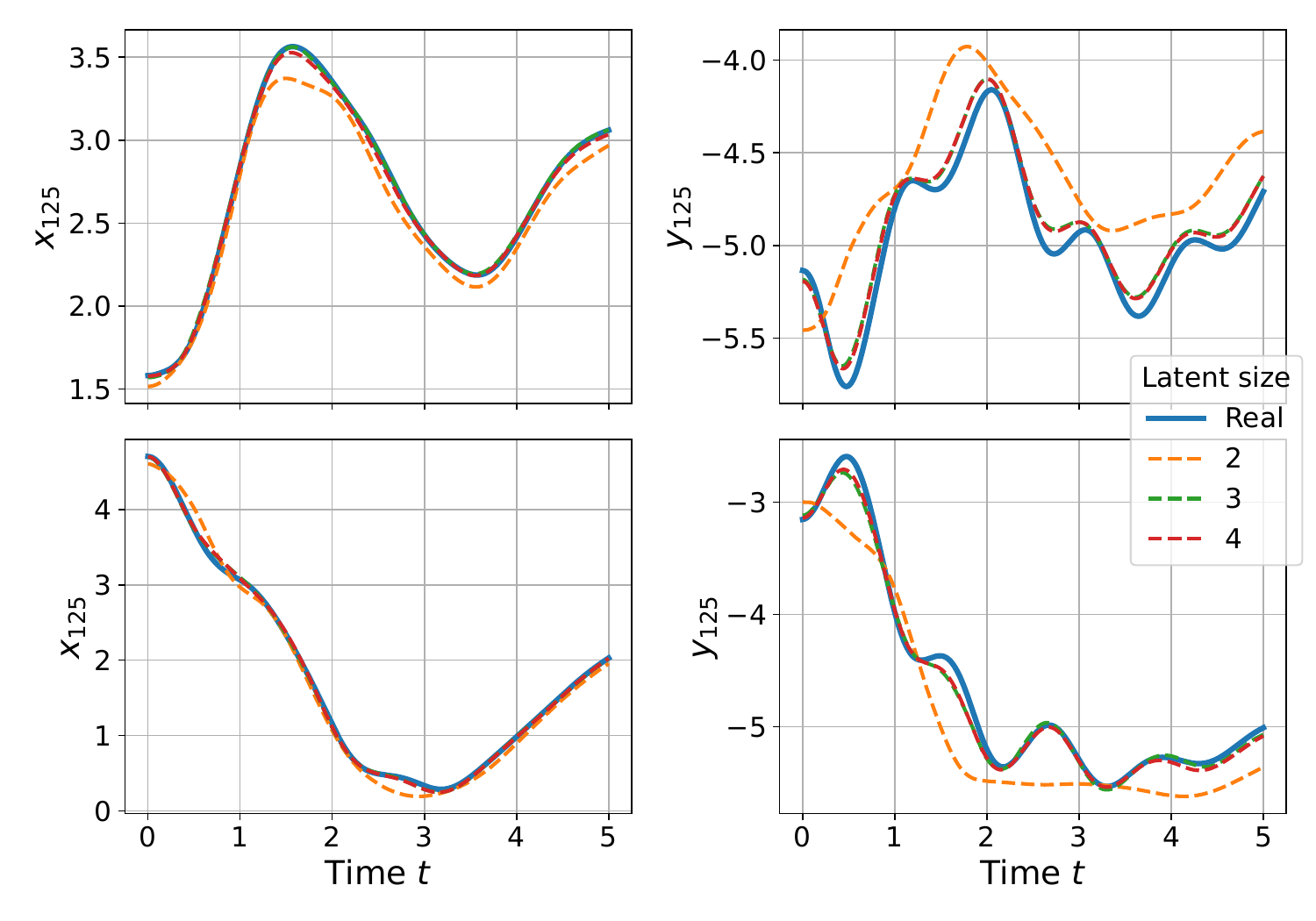}
    \caption{\small Example of reconstructed trajectory with different sizes of latent state.}
    \label{fig:gap-oscill}
\end{figure}

We further analyze how much the approach can compress the systems' state. Figure \ref{fig:compression} shows the test MSE on the five systems varying the latent state size for both flat and graph autoencoders. In most cases, 3 variables are enough to efficiently represent the system's state with a reasonable approximation error, while using more variables typically results in small or marginal improvements in the error. Using 2 variables seems to not be enough to effectively capture all systems' behaviours as can also be observed from the reconstructed trajectory of some of the masses in Figure \ref{fig:gap-oscill}. We can notice that, with 2 variables, minor oscillations are not correctly reconstructed, and there is a consistent gap between the real and reconstructed trajectory. This does not happen when using 3 or more variables.
The 2 variables case is also useful to show what happens when the assumption that the model achieves close-to-zero loss does not hold. Indeed, the reduced model still approximate the real-space model, although losing the ability to represent some of its particularity. We also believe the models can be successfully used for latent space simulations, although these could deviate much more from the real ones.

\subsection{Control experiments}

\paragraph{Setup}
We test the proposed controller by simulating its application for planar posture regulation on the second spring network in Figure \ref{fig:systems}.
The system is actuated by a generalized force $\tau \in \mathbb{R}^2$ applied to a single mass at each simulation step.
Our controller employs an autoencoder with latent size $m=2$, trained as in the previous sections. 
The actuation matrix is therefore the selection matrix 
    $\big[\underbrace{0\cdots0}_{a-1}\ I\ \underbrace{0\cdots0}_{N-a}\big] \in \mathbb{R}^{2\times2n}$,
%
where $a$ is the index of the actuated mass and $I\in\mathbb{R}^{2\times2}$ is the identity matrix.

%
We perform 50 simulations, randomly selecting the target configuration among the configurations $\bar{q}$ in the training/validation simulations. The actuation mass is randomly selected among three fixed candidates chosen in correspondence with the lateral structures and as far as possible from the structure edges. 
The initial state is always the rest position with zero initial velocity. The simulations are 5 seconds long. 

\begin{figure}
    \centering
    \includegraphics[trim = {0 10 0 0 },clip,width=.95\linewidth]{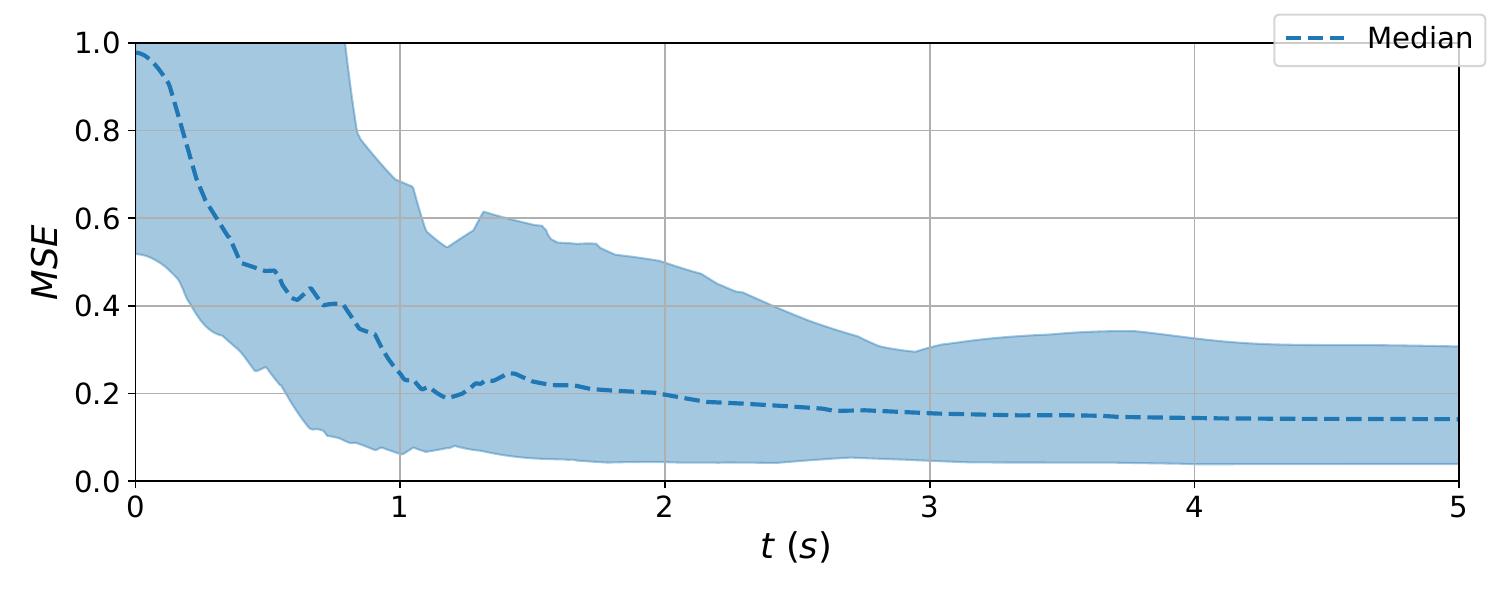}
    \caption{\small Evolution of the normalized MSE through time and across several simulation, calculated between the reference configuration $\bar{q}$ and the actual configuration $q(t)$. The shaded are refers to the first and third quartile.}
    \label{fig:control-error}
\end{figure}

\paragraph{Results}
We evaluate the controller according to the MSE between the full-space system configuration at time $t$ and the target configuration $\bar{x}$. We report the resulting Figure \ref{fig:control-error}. 
The dashed line represents the median error among the considered simulations, while the band represents the 25-75 percentiles of the error.

\section{Conclusions}
This work investigated the application of deep autoencoders to the compression of high-dimensional dynamical systems, while maintaining Hamiltonian/Lagrangian structural properties in the low-dimensional approximation.
The approach was extensively validated and evaluated over several high-dimensional mass-spring-damper models. The reduced systems were exploited to perform simulation in the latent space, from which the original complete space evolution were reconstructed. We also proposed a possible usage of such compressed representations for planar posture regulation of highly underactuated systems, evaluating the developed controller in simulations.
%
%
Future work will focus on developing and testing data-driven control algorithms for high-dimensional systems where model-based strategy are used in conjunction with the learned latent model. 
%
%
We will also dive into extending the approach to systems whose Hamiltonian differs from \eqref{eq:energy}, including fluid dynamics \cite{maulik2021reduced} and astronomy \cite{gabbard2022bayesian}.

\bibliographystyle{myIEEEtran}
\balance
\bibliography{references}

\appendix

\subsection{Data Generation} \label{app:data-generation}
We detail here the generation of our dataset. We create five high-dimensional systems by randomly chaining triangular and square meshes. The generation algorithm creates a base mesh chain while at the same time randomly attaching lateral structures in order to increase the variability of the generated models. Each system has a total of 200 masses (i.e., $n = 400$ degrees of freedom) and $e=636$ connections. The configuration vector $q$ is build such that the $2i\--$th and $(2i+1)\--$th elements $q_{2i},q_{2i + 1}$ are the $x$ and $y$ positions of the $i\--$th mass. The resulting inertial matrix $M \in \mathbb{R}^{400 \times 400}$ in equation (3) is diagonal and its entries are the masses associated with each degree of freedom $m_i$.

The potential energy in equation (3) is
\begin{equation}\small \label{eq:system_potential}
    V(q) = \sum_{i=1}^{\frac{n}{2}} m_i g (q_{2i}\sin(\theta)+q_{2i + 1}\cos(\theta)) + \frac{1}{2}\sum_{j=1}^{e}k_i(l_j(q) - l_j^0)^2.
\end{equation}
The first term in \eqref{eq:system_potential} is the gravitational energy associated to all the masses. We consider systems immersed in a constant gravitational field, with intensity $g\in\mathbb{R}$ and orientation angle $\theta \in [0,2\pi)$. The second term in \eqref{eq:system_potential} is the elastic energy associated to all the connections, with $k_i$ spring stiffness, $l_j$ current spring length, and $l_j^0$ spring rest length. Note that $l_j$ is a nonlinear function of the positions of the two masses that the spring connects. We assume a damping element to be in parallel to each spring, and the damping matrix $D(q) \in \mathbb{R}^{400 \times 400}$ in equation (3) is calculated by differentiating all $l_j$ w.r.t. $q$.

The gravity conditions for the 7 training simulations are manually chosen in order to include into the training data configurations as diverse as possible, while covering the configuration space as much as possible. In particular, as gravity intensity we selected $g \in \{9.81, \dots, 9.81, 6, 14\} \frac{\mathrm{m}}{\mathrm{s}^2}$  while the as orientation angle we selected $\theta \in \{-3/4\pi, -\pi/3, -\pi/4, -2/3\pi, \pi/2, -\pi/3 e -2/3\pi\}$. The gravity conditions for the 28 test simulations are instead randomly selected from two uniform distributions such that $g\in[3, 17]\ \frac{\mathrm{m}}{\mathrm{s}^2}$ and $\theta\in[-\frac{3}{4}\pi, -\frac{1}{4}\pi]$.

In some cases, the simulation of a system could produce a very static trajectory, meaning that the resulting configurations are very similar to each other. This could make the data biased towards certain configurations that are much more frequent than others, since having multiple similar examples means giving them more importance. This would, in turn, produce models biased towards reconstructing the most frequent configurations. Therefore, for each simulation, the resulting configurations are filtered in order to remove those too that are too similar. In particular, all the selected configurations, we ensure the data does not contain any other configuration at (Euclidean) distance lower than $\epsilon$ around it. In the experiments, the value for $\epsilon$ is always set to $0.1$.

\begin{table*}[t!]
    \centering
    \begin{tabular}{c|c|c|c|c|c|c|c}
        System & Model & $\sigma=0.01$ & $\sigma=0.05$ & $\sigma=0.1$ & $\sigma=0.5$ & $\sigma=1$ & $\sigma=5$ \\
        \hline
        \multirow{2}{*}{1} & Flat AE & $3.00\mathrm{e}{-4}$ & $3.52\mathrm{e}{-4}$ & $5.06\mathrm{e}{-4}$ & $5.57\mathrm{e}{-3}$ & $2.07\mathrm{e}{-2}$ & $4.12\mathrm{e}{-1}$ \\
                          & Graph AE & $3.95\mathrm{e}{-4}$ & $5.11\mathrm{e}{-4}$ & $8.94\mathrm{e}{-4}$ & $1.40\mathrm{e}{-2}$ & $6.62\mathrm{e}{-2}$ & $1.03$ \\
        \hline
        \multirow{2}{*}{2} & Flat AE & $1.57\mathrm{e}{-4}$ & $1.90\mathrm{e}{-4}$ & $2.97\mathrm{e}{-4}$ & $3.73\mathrm{e}{-3}$ & $1.40\mathrm{e}{-2}$ & $2.60\mathrm{e}{-1}$ \\
                          & Graph AE & $1.33\mathrm{e}{-4}$ & $1.39\mathrm{e}{-4}$ & $1.58\mathrm{e}{-4}$ & $7.89\mathrm{e}{-3}$ & $2.94\mathrm{e}{-2}$ & $1.01$ \\
        \hline
        \multirow{2}{*}{3} & Flat AE & $2.58\mathrm{e}{-4}$ & $2.92\mathrm{e}{-4}$ & $3.99\mathrm{e}{-4}$ & $3.91\mathrm{e}{-3}$ & $1.46\mathrm{e}{-2}$ & $2.72\mathrm{e}{-1}$ \\
                          & Graph AE & $5.24\mathrm{e}{-4}$ & $5.95\mathrm{e}{-4}$ & $8.09\mathrm{e}{-4}$ & $7.75\mathrm{e}{-3}$ & $3.12\mathrm{e}{-2}$ & $1.16$ \\
        \hline
        \multirow{2}{*}{4} & Flat AE & $8.90\mathrm{e}{-4}$ & $9.22\mathrm{e}{-4}$ & $1.02\mathrm{e}{-3}$ & $4.10\mathrm{e}{-3}$ & $1.36\mathrm{e}{-2}$ & $2.61\mathrm{e}{-1}$ \\
                          & Graph AE & $8.34\mathrm{e}{-4}$ & $9.32\mathrm{e}{-4}$ & $1.25\mathrm{e}{-3}$ & $1.16\mathrm{e}{-2}$ & $4.53\mathrm{e}{-2}$ & $9.46\mathrm{e}{-1}$ \\
        \hline
        \multirow{2}{*}{5} & Flat AE & $9.54\mathrm{e}{-4}$ & $9.89\mathrm{e}{-4}$ & $1.10\mathrm{e}{-3}$ & $4.67\mathrm{e}{-3}$ & $1.52\mathrm{e}{-2}$ & $2.86\mathrm{e}{-1}$ \\
                          & Graph AE & $8.13\mathrm{e}{-4}$ & $8.47\mathrm{e}{-4}$ & $9.53\mathrm{e}{-4}$ & $4.69\mathrm{e}{-3}$ & $1.88\mathrm{e}{-2}$ & $6.84\mathrm{e}{-1}$ \\
    \end{tabular}
    \caption{Average MSE between real and reconstructed configuration using noisy inputs for reconstruction. The input configurations $q$ are altered by adding normal noise with varying level of standard deviation $\sigma$.}
    \label{tab:noise-score}
\end{table*}

\subsection{Training and Model Selection} \label{app:models}
A flat autoencoder is trained for each system, using the data coming from its simulations. The architecture of the autoencoder is the same for all the systems and consists of 4 encoding layers and 4 decoding layers. The encoder's layers sizes are 300, 200, 100 and 5 respectively. Hence, the latent size is set to 5, selected to be around 1\% of the real systems size. The decoder's layer sizes are 100, 200, 300 and 400 respectively, resulting in $\sim400$K parameters in total. The \texttt{ELU} activation function is used in the first three encoding layers and in the first three decoding layers, while the last layer of the encoder and the decoder use a linear activation function.

Also a graph autoencoder was trained for each system. We set all the MLPs as single linear projections so that all the non-linear processing is preformed by the graph convolutional network. Both the encoding and decoding GCNs are made of 4 layers with 50 node channels each. Each layer contains the convolutional operator followed by an \texttt{ELU} non-linearity. The encoding node-wise projection have 10 output units, while the encoding graph-level projection have $10\times200=2000$ input units and 5 output units. Thus, the latent state size is still set to five. The latent-to-node states projection have 2000 output units, resulting in 10 input channels per node in the decoding GCN, while the decoding node-wise projection has two output units. This results in a total of about 54K parameters.

All the considered models are trained for 500 epochs using Adam as optimization algorithm. All the algorithm's parameters are left to their default values, except for the learning rate. In addition, L2 regularization is used to prevent overfitting.

For both the types of architecture, preliminary experiments showed how the training is particularly sensitive to the learning rate value. Indeed, using a fixed learning rate for the entire training resulted in either an extremely unstable training/validation score or in a too slow convergence. In order to mitigate this problem, we resort to a variable learning rate approach, where we adjust the learning rate during the training. In particular, we use a step policy for the learning rate adjustment: every \texttt{step} epochs the learning rate is updated according to the multiplicative factor $\gamma$:
\begin{equation*}
    \texttt{lr}_{e+1} = \begin{cases} \texttt{lr}_e * \gamma & \text{if}\ e\ \text{mod}\ \texttt{step} = 0 \\ \texttt{lr}_e & \text{otherwise}\end{cases}
\end{equation*}
where $e$ is the current epoch and \texttt{lr} is the learning rate. For values of $\gamma < 0$, this results in a decaying learning rate, allowing an higher value in the early epochs while at the same time stabilizing the training in the later epochs.

We note here that this behavior is unusual for algorithms like Adam, which already use an adaptive learning rate, and could be caused by the unnormalized data or by the low error reached in the later epochs. Nevertheless, the unstable behavior made it harder to obtain a statistically valid evaluation of the models. Thus, the decision to resort to a variable learning rate.

\begin{table}[ht]
    \centering
    \begin{tabular}{c|c|c}
    \hline
         & Flat Autoencoder & Graph Autoencoder \\
        \hline
        $\texttt{lr}$   & $1\mathrm{e}{-3}$, $5\mathrm{e}{-4}$, $1\mathrm{e}{-4}$ & $1\mathrm{e}{-2}$, $6\mathrm{e}{-3}$, $3\mathrm{e}{-3}$, $1\mathrm{e}{-3}$ \\
        $\lambda$       & $1\mathrm{e}{-5}$, $1\mathrm{e}{-6}$, $1\mathrm{e}{-7}$ & $1\mathrm{e}{-5}$, $1\mathrm{e}{-6}$, $1\mathrm{e}{-7}$ \\
        $\gamma$        & 0.3, 0.5 & 0.7, 0.8 \\
        $\texttt{step}$ & 100, 200 & 30, 40 \\
        \hline
    \end{tabular}
    \caption{Hyperparameter values used in the grid search for the flat autoencoder and graph autoencoder.}
    \label{tab:gs-values}
\end{table}
For each model and for each system, a grid search is performed to select the best training hyperparameters. In particular, the considered hyperparameters are: the learning rate $\texttt{lr}$, the L2 weight decay $\lambda$, the learning rate decay factor $\gamma$ and decay step \texttt{step}. Table \ref{tab:gs-values} reports the hyperparameters values for the two models. These are the same for all the considered systems.

\subsection{Noisy Data Experiments}
Since the availability of noise-free measurements is usually extremely limited in real-world scenarios, we assessed how robust the models are against noisy measurements. This scenario was simulated by adding to each real system configuration $q$ a noise vector $\epsilon$ drawn from a normal distribution $\mathcal{N}(0, \sigma^2)$. The resulting configuration $q_{\epsilon}$ is then fed into the autoencoder models for reconstruction, obtaining a reconstructed configuration $\tilde{q}_{\epsilon}$. The reconstruction is then matched against the original noise-free configuration using the MSE:
\begin{equation}
    \mathcal{L}_{\mathrm{noise}}(q) = ||q - \tilde{q}_{\epsilon}||_2^2 = ||q - \mathcal{D}(\mathcal{E}(q_{\epsilon}))||_2^2
\end{equation}
Different value for the standard deviation $\sigma$ were tested, varying from $0.01$ to $5$. Table \ref{tab:noise-score} shows the average resulting scores for all the systems and models, and values for $\sigma$. We can notice that, for a magnitude of the standard deviation around $0.1$ (corresponding to 1 decimeter) the obtained scores remain in the same order of magnitude as in the noiseless case. The error then quickly increases for standard deviations in the order of $0.5$ and $1$ (i.e., around half of a meter and a meter).

\subsection{Latent Variables Interpretability}
It is interesting to investigate the role of the learnt latent variables, to see if they have a particular physical meaning, or control specific characteristic of the corresponding systems.

\begin{figure*}
    \centering
    \includegraphics[width=0.45\linewidth]{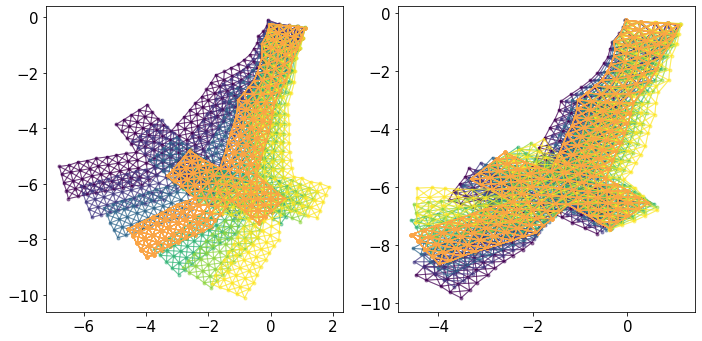}
    \includegraphics[width=0.45\linewidth]{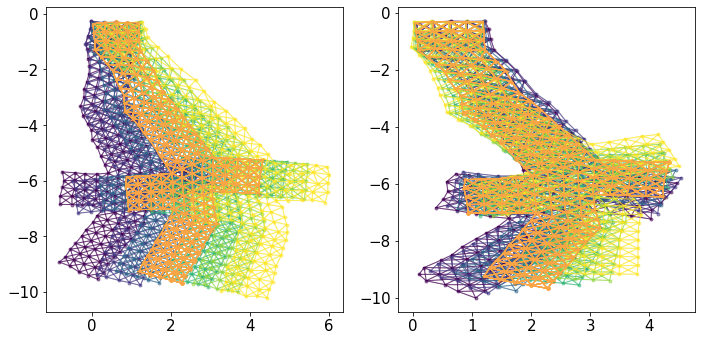}
    \includegraphics[width=0.45\linewidth]{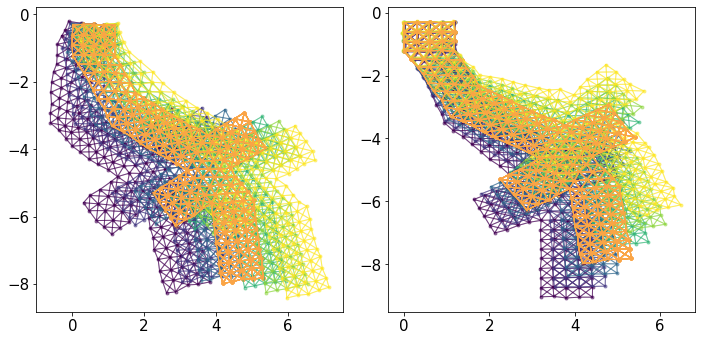}
    \includegraphics[width=0.45\linewidth]{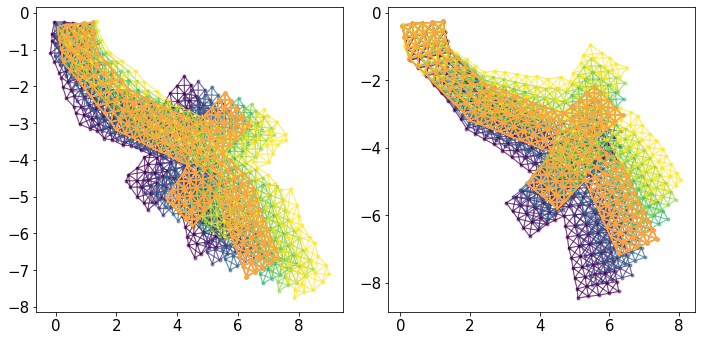}
    \caption{Effect of altering one latent variable at a time in a flat autoencoder with latent size 2 trained on system 1}
    \label{fig:variables-sys1}
\end{figure*}

\begin{figure*}
    \centering
    \includegraphics[width=0.45\linewidth]{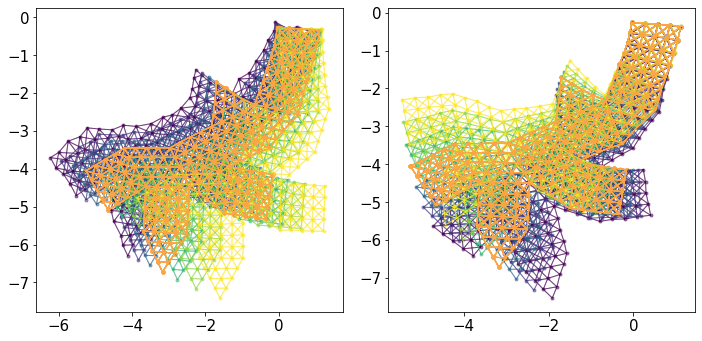}
    \includegraphics[width=0.45\linewidth]{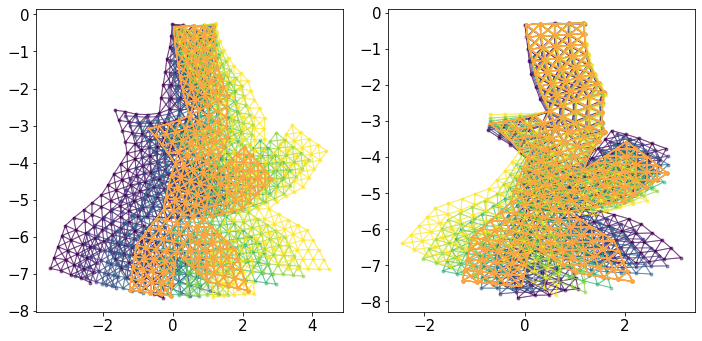}
    \includegraphics[width=0.45\linewidth]{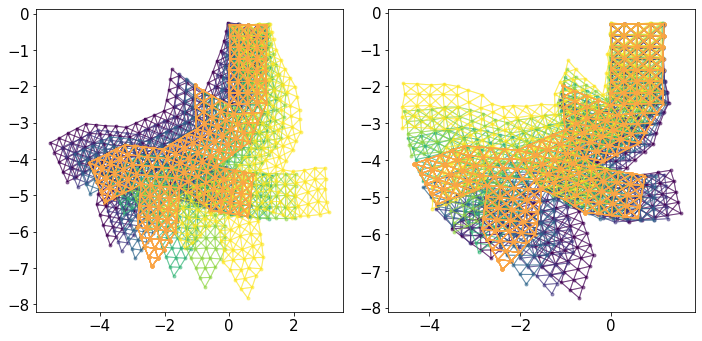}
    \includegraphics[width=0.45\linewidth]{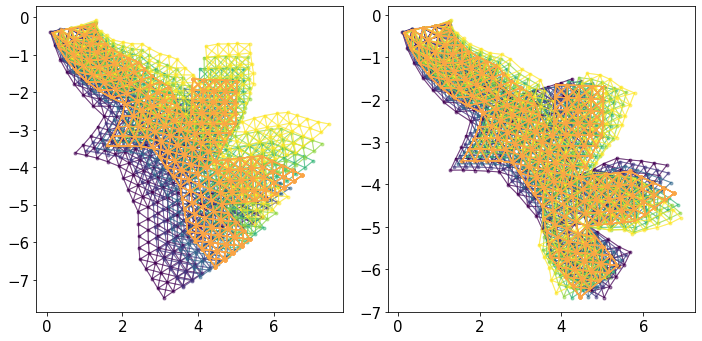}
    \caption{Effect of altering one latent variable at a time in a flat autoencoder with latent size 2 trained on system 2}
    \label{fig:variables-sys2}
\end{figure*}

\begin{figure*}
    \centering
    \includegraphics[width=0.45\linewidth]{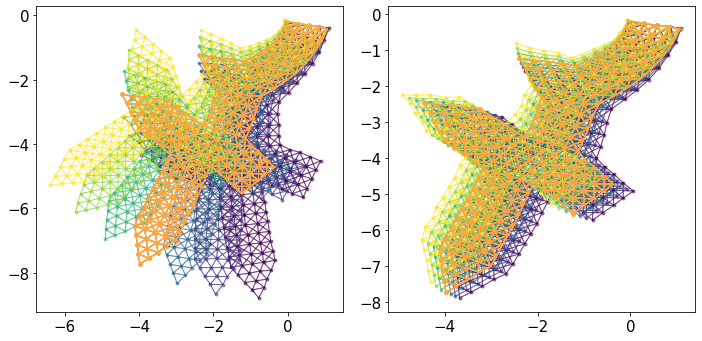}
    \includegraphics[width=0.45\linewidth]{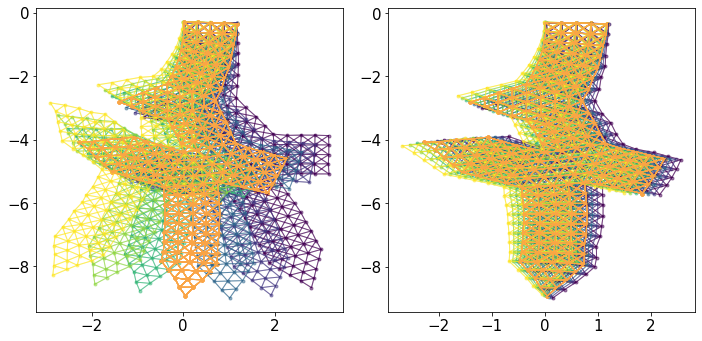}
    \includegraphics[width=0.45\linewidth]{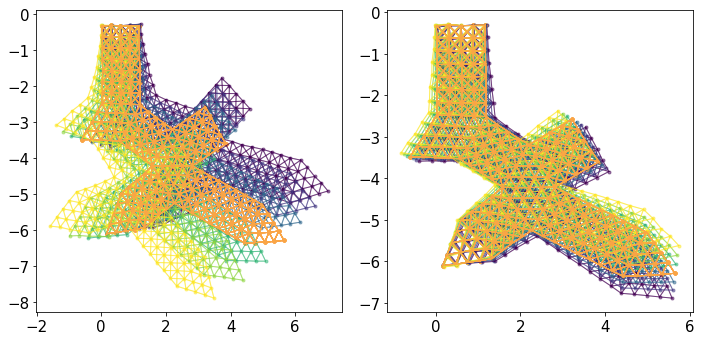}
    \includegraphics[width=0.45\linewidth]{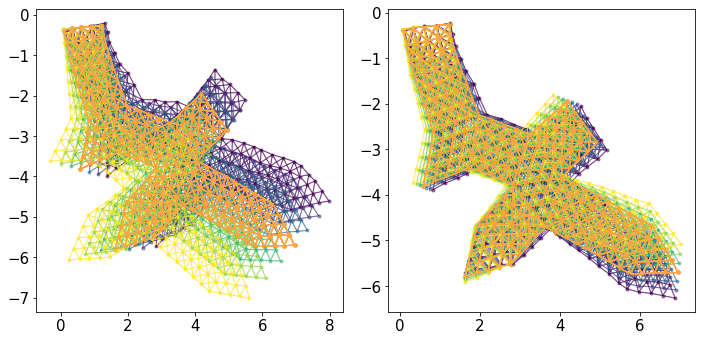}
    \caption{Effect of altering one latent variable at a time in a flat autoencoder with latent size 2 trained on system 3}
    \label{fig:variables-sys3}
\end{figure*}

\begin{figure*}
    \centering
    \includegraphics[width=0.45\linewidth]{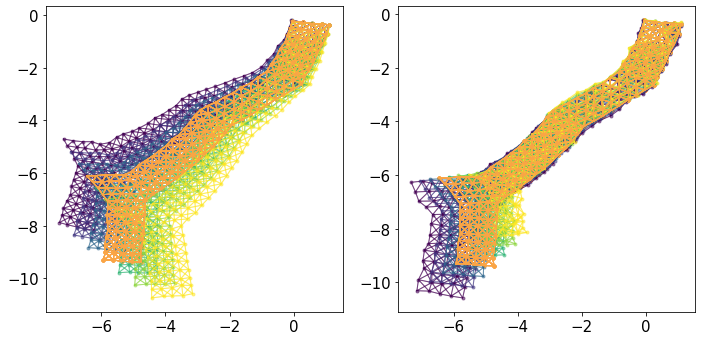}
    \includegraphics[width=0.45\linewidth]{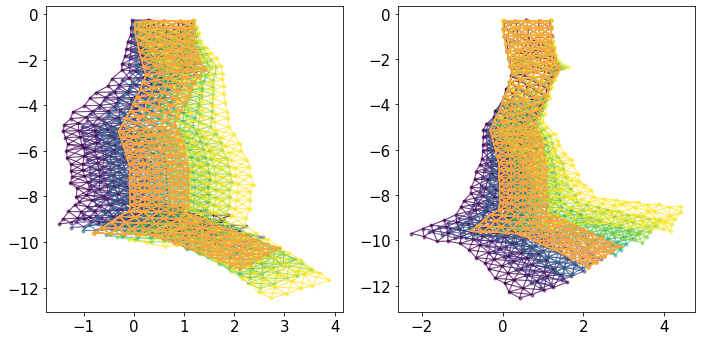}
    \includegraphics[width=0.45\linewidth]{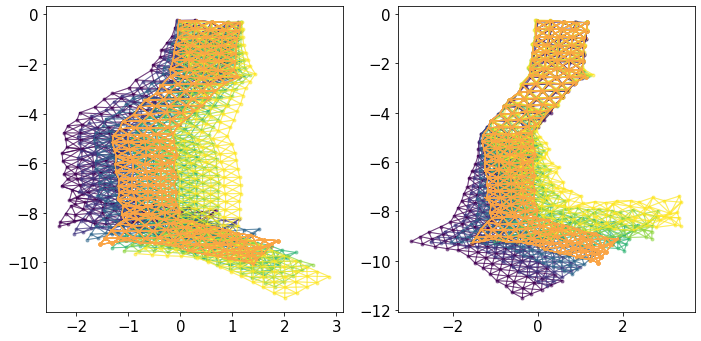}
    \includegraphics[width=0.45\linewidth]{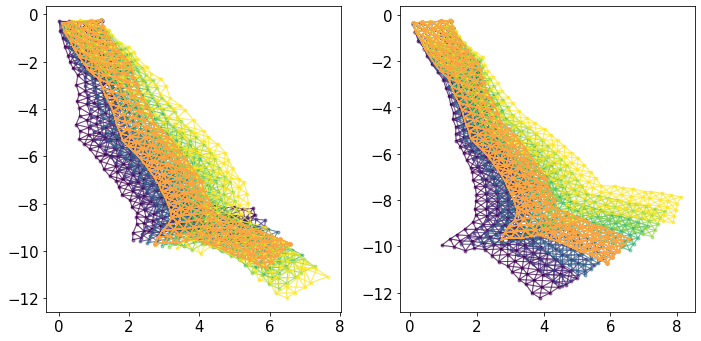}
    \caption{Effect of altering one latent variable at a time in a flat autoencoder with latent size 2 trained on system 4}
    \label{fig:variables-sys4}
\end{figure*}

\begin{figure*}
    \centering
    \includegraphics[width=0.45\linewidth]{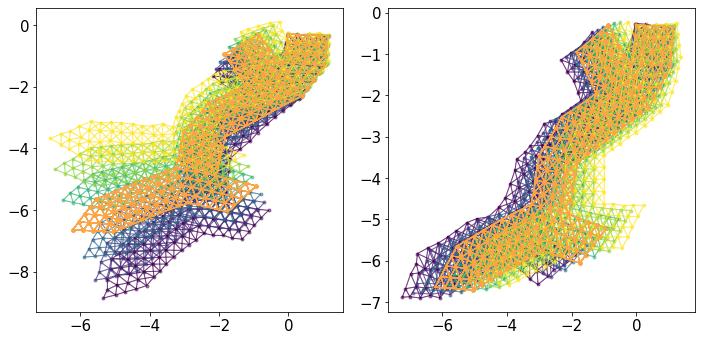}
    \includegraphics[width=0.45\linewidth]{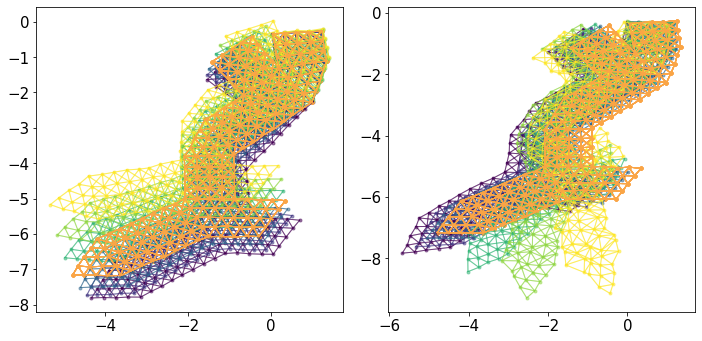}
    \includegraphics[width=0.45\linewidth]{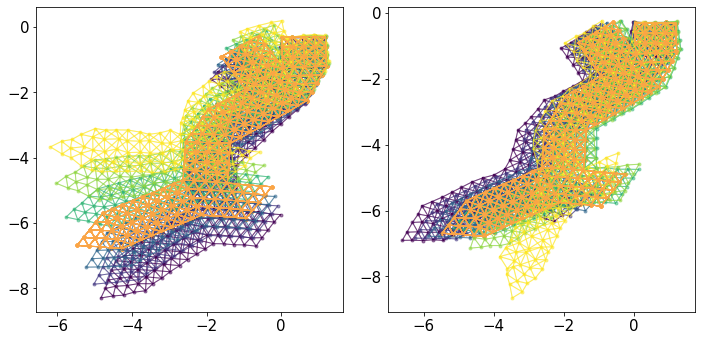}
    \includegraphics[width=0.45\linewidth]{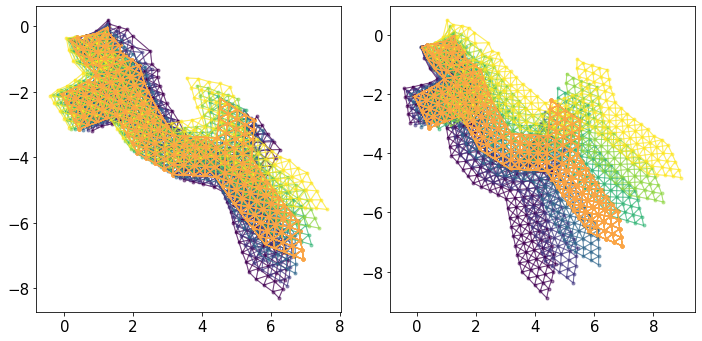}
    \caption{Effect of altering one latent variable at a time in a flat autoencoder with latent size 2 trained on system 5}
    \label{fig:variables-sys5}
\end{figure*}

In order to do so, we selected particular real space configurations for each system. In particular we selected four configurations, corresponding to: 1. the rest position; 2,3. positions where the system is leaning towards the left and towards the right; and 4. a position where system is extended downwards.
For each of the selected configurations $q$, we computed its latent representation $\xi^q = \mathcal{E}(q)$ and then we independently altered the value of each latent variable $i$, obtaining a new latent representation $\xi^q_i$ which differs from the original $\xi^q$ one only for variable $i$. We tested different magnitude of change in the variables' value. In particular, each variable was altered by the 7\%, 14\% and 21\% of its total range (computed across all available simulations), in both direction, resulting in 6 new latent vectors.
We then used the newly computed latent representation to obtain the corresponding real space configuration $q_i$ by feeding it to the decoder network $q_i = \mathcal{D}(\xi^q_i)$.
Figures \ref{fig:variables-sys1}-\ref{fig:variables-sys5} show the original configuration $q$ (in orange) against the obtained configurations. Each sub-figure shows the reference configuration against the 6 configurations obtained by altering a single variable. The considered models are flat autoencoders with 2 latent variables. Similar results were obtained with a graph autoencoder model.

We notice how the variables tend to control particular structural aspect of the systems, such as the overall rotation of the structure or the relative rotation of sub-structures of the systems. We also note here that the configurations reconstructed from the altered latent states tend to be more unstable as the magnitude of the alteration increases. This shows that the two variables are not in general completely independent, as a totally independent assignment of values to the variables usually produces messy reconstructed configurations. This aspect could be avoided by regularizing the latent space, for instance via variational approaches, in order to disentangle the latent variables, obtaining models that are more interpretable and steerable than the simpler ones.

\end{document}